\documentclass[aps, prx, twocolumn, superscriptaddress, nofootinbib, floatfix]{revtex4-2}

\usepackage{graphicx}  % For figures
\usepackage{dcolumn}   % Align table columns on decimal point
\usepackage{bm}        % Bold math
\usepackage{amsmath}   % Math environments
\usepackage{amssymb}   % Math symbols
\usepackage{hyperref}  % Hyperlinks
\usepackage{xcolor}    % Colors for comments/drafting
\usepackage{booktabs}  % Professional table formatting

\hypersetup{
    colorlinks=true,
    linkcolor=blue,
    citecolor=blue,
    urlcolor=blue
}

\begin{document}

% --- Front Matter ---

\title{OSOG: A Differentiable, Physics-Informed Synthetic Data Engine \\ for Micro-Optical Environments}

\author{Caio Silva}
\affiliation{Department of Physics, Massachusetts Institute of Technology, Cambridge, Massachusetts 02139, USA}
 \email{caiosiq@mit.edu}

\date{\today}

\begin{abstract}
Deep learning in computational microscopy is severely constrained by the scarcity of densely annotated datasets. While synthetic data generation has bridged this gap in macroscopic computer vision, traditional graphics engines rely on geometric ray-tracing, failing to capture the micro-optical phenomena required for microscopy. Conversely, while wave-optics formulations exist, rendering them computationally tractable at the scale required for deep learning remains a massive systems challenge. To address this, we introduce the Optical Synthetic Object Generator (OSOG), a high-performance, fully differentiable forward-modeling engine. Drawing on established physical models of diffraction and phase retardation, OSOG maps continuous Optical Path Difference (OPD) calculations into a highly optimized, PyTorch-native Structure-of-Arrays (SoA) architecture. We validate this computational framework across three axes: First, object detection models (YOLOv11-OBB) trained purely on OSOG-generated data achieve robust zero-shot transfer to real-world highly occluded Lysozyme micrographs. Second, we introduce DiffOSOG, demonstrating that the engine's end-to-end differentiability allows for the exact recovery of continuous optical parameters via curriculum-guided inverse rendering. Finally, OSOG bypasses the $\mathcal{O}(N)$ bottlenecks of sequential ray-tracing, demonstrating sub-linear scaling by synthesizing 40,000 complex wave-optic particles in under 50 milliseconds (\>20 FPS). By providing a fast, scalable, and physically grounded tensor pipeline, OSOG enables true real-time, on-the-fly dataset generation.
\end{abstract}

\maketitle

% --- Main Body ---

\section{Introduction}

Deep learning has fundamentally transformed computational microscopy, enabling rapid advancements in automated cell tracking, defect detection in materials science, and in-situ crystallization monitoring. However, the performance of these models—from standard U-Nets to modern Vision Transformers—is intrinsically bounded by the quality and volume of their training data. In microscopy, acquiring large-scale, densely annotated datasets is notoriously difficult. Human annotation of dense, overlapping suspensions is not only labor-intensive but often physically impossible when dealing with transparent or semi-transparent objects under modalities like Differential Interference Contrast (DIC) or Oblique Illumination.

In macroscopic computer vision (e.g., autonomous driving, robotics), this data bottleneck is routinely bypassed using synthetic datasets generated by commercial 3D engines like Blender or Unreal Engine \cite{tobin2017domain, tremblay2018training}. Unfortunately, applying this paradigm to microscopy introduces a severe ``Reality Gap.'' Traditional rendering engines are built on the principles of geometric optics—simulating light as discrete rays traveling in straight lines.

% \begin{figure}[t]
%     \centering
%     \includegraphics[width=\linewidth]{figures/Figure 1.pdf}
% \caption{\textbf{Conceptual Motivation: Geometric Optics vs. Wave Optics.} \textbf{(a)} The fundamental difference in light transport paradigms. Traditional synthetic data engines rely on geometric optics, modeling light as macroscopic rays and ignoring wave interference. In contrast, the micro-optical domain is governed by wave optics, where light undergoes phase retardation and diffraction. \textbf{(b)} A visual demonstration of how rendering engines utilizing each form of optics produce inherently different outputs. Standard ray-tracing fails to capture micro-optical phenomena, yielding flat representations, whereas the OSOG wave-optics pipeline natively reproduces the complex directional shadowing, optical shear, and diffraction fringes required for accurate microscopy simulation.}
%     \label{fig:ray_vs_wave}
% \end{figure}

While ray tracing excels at macroscopic shadows and reflections, it fundamentally fails at the microscopic scale, where the wave nature of light dominates. At this scale, image formation is governed by diffraction limits (Airy disks), phase interference, and structural waveguiding. Consequently, datasets generated via standard geometric rendering fail to capture the Optical Path Difference (OPD) and phase gradients necessary for realistic microscopy simulation, rendering them largely ineffective for Sim-to-Real model transfer.

To solve this, we introduce the Optical Synthetic Object Generator (OSOG): a high-performance, fully differentiable forward-modeling engine \cite{kato2018neural, mildenhall2020nerf} designed from the ground up for wave-propagation microscopy. Built entirely in PyTorch, OSOG discards the standard ray-tracing pipeline in favor of calculating exact phase shifts and modulating complex wavefronts through a virtual microscope objective. Furthermore, to address the computational bottlenecks of traditional object-oriented simulators, OSOG employs a GPU-native Structure-of-Arrays (SoA) architecture. By vectorizing physical properties and utilizing a novel scatter-add composition technique, OSOG can render thousands of procedurally generated, optically complex particles in milliseconds without the data leaving the GPU VRAM.

Because every operation within OSOG—from 3D spatial rotation to sensor noise application—is differentiable, it serves as a true ``Digital Twin'' for optical systems. This opens the door not only to generating infinite, pixel-perfect ground truth data for robust neural network training, but also to solving inverse-design problems directly from empirical data.

\subsection{Principal Contributions}
Our core contributions are threefold:
\begin{enumerate}
    \item \textbf{A PyTorch-Native SoA Rendering Architecture:} We implement established wave-optics phase formulations into a highly optimized, fully vectorized tensor pipeline, bypassing the computational bottlenecks of traditional object-instanced ray-tracing.
    \item \textbf{Differentiable Domain Calibration (DiffOSOG):} We formulate the engine natively in PyTorch to support continuous gradient flow, demonstrating exact optical parameter recovery (solving the inverse rendering problem) via a composite semantic loss.
    \item \textbf{Zero-Shot Sim-to-Real Transfer:} We empirically validate the engine's output fidelity by training a YOLOv11-OBB model entirely on uncurated synthetic data, achieving robust zero-shot inference on real-world, heavily occluded biological microscopy targets.
\end{enumerate}

\section{Related Works}

\subsection{The Limits of Geometric Optics}
The generation of synthetic datasets for computer vision has traditionally relied on commercial rendering engines such as Blender, Unity, or Unreal Engine. While highly effective for macroscopic scenes, these engines fundamentally rely on geometric ray tracing. At the microscopic scale, light behavior is dominated by wave phenomena—such as diffraction, interference, and phase shifts—which ray tracing cannot natively resolve. Consequently, synthetic microscopy datasets generated via standard computer graphics often fail to accurately model the depth-dependent Point Spread Functions (PSF), the Airy disks imposed by the diffraction limit, or phase-contrast modalities like DIC.

\subsection{Differentiable Wave Optics and Forward Modeling}
The computational imaging community has recently recognized the necessity of explicitly integrating physical forward models into deep learning pipelines. Bouchama et al. (2023) \cite{bouchama2023deep} demonstrated that incorporating the exact microscope image formation model into a neural network enables highly efficient reconstruction, even in low-data regimes. Similarly, Sun et al. (2024) \cite{sun2024hybrid} highlight the superiority of hybrid frameworks that combine deep learning with physics-based neural networks for computational microscopy.

Recent works have introduced differentiable optical simulators to facilitate this. \textit{Chromatix} \cite{deb2024chromatix} provides a JAX-based framework for GPU-accelerated wave optics, emphasizing accurate PSF modeling. Concurrently, \textit{deltaMic} \cite{ichbiah2025deltamic} introduced a 3D differentiable renderer specifically tailored for fluorescence microscopy. OSOG builds upon this paradigm but fundamentally diverges in architecture and scope, introducing a highly scalable SoA pipeline capable of rendering thousands of procedurally generated, multi-phase particles across multiple optical modalities simultaneously.

\subsection{Bridging the Sim-to-Real Gap}
The transition toward machine-learning-powered workflows in microscopy is heavily bottlenecked by data acquisition. As Morgado et al. (2024) \cite{morgado2024rise} note, the requirement for large, pixel-perfect annotated datasets remains a massive challenge, particularly for multi-modal imaging or dense suspensions. Recent successes in industrial imaging have proven that the Sim-to-Real domain gap can be bridged if the synthetic data accurately reflects the underlying physics. OSOG directly translates this methodology to microscopy, serving as a comprehensive digital twin for zero-shot Sim-to-Real transfer.

\section{Methodology}

The OSOG framework is mathematically formulated as an end-to-end, fully differentiable forward model $f(\Theta) = \hat{I}$. Here, $\Theta$ denotes a stochastic state vector containing the physical parameters of the simulated environment (e.g., 3D coordinates, morphological degrees of freedom, refractive indices, and continuous optical parameters), and $\hat{I}$ is the synthesized multi-modal micrograph. 

To bypass the severe computational bottlenecks of traditional sequential rendering, OSOG decomposes the image formation process into a highly vectorized, four-stage pipeline. First, the engine transforms the state vector $\Theta$ into a unified Geometric Buffer (G-Buffer) representing the physical sample geometry. Second, an analytic projection computes the complex exit wavefront $U(x,y)$ emerging from the sample. Third, this wavefront is modulated by the specific transfer functions of the virtual microscope's optical head (simulating the objective lens, numerical aperture, and illumination modalities). Finally, the modulated patches are composited onto a global sensor array using zero-cost masked splatting, where rigorous physical regularization (e.g., optical aberrations, slide fouling, and sensor noise) is applied to produce the final digital output. 

\begin{figure*}[t]
    \centering
        \includegraphics[width=\textwidth]{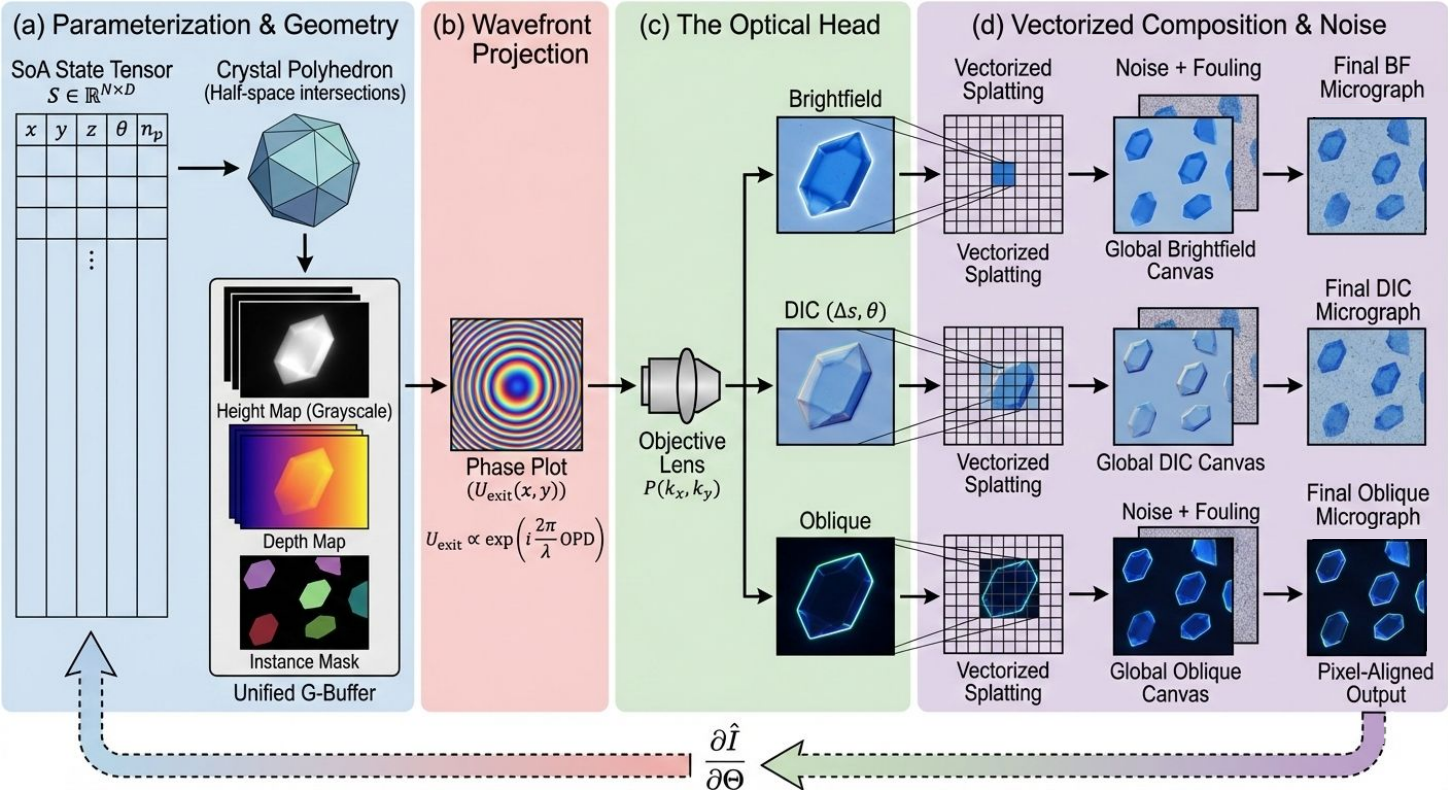}

    \caption{\textbf{The OSOG Differentiable Forward Pipeline.} \textbf{(a) Parameterization:} The SoA state tensor drives procedural half-space intersections to generate the Unified G-Buffer (depth, height, masks). \textbf{(b) Wavefront Projection:} The geometry is translated into an Optical Path Difference (OPD) to construct the complex exit wave. \textbf{(c) Optical Head Modalities:} The wavefront is modulated by the pupil function to simulate Brightfield, DIC, and Oblique contrast natively. \textbf{(d) Sensor Composition:} Vectorized splatting composites the localized optical patches onto a global canvas, followed by the application of physical artifacts (slide fouling, sensor noise) for robust domain regularization.}
    \label{fig:pipeline_full}
\end{figure*}

\subsection{Procedural Geometry and the Unified G-Buffer}
To circumvent the $O(N)$ scaling limitations of traditional object-oriented particle simulation, OSOG is architected around a Structure-of-Arrays (SoA) paradigm. The physical state of $N$ particles within a field of view is parameterized as a dense tensor $\mathcal{S} \in \mathbb{R}^{N \times D}$.

The macroscopic morphology of the target suspensions is generated via a differentiable half-space intersection algorithm. A specific euhedral crystal habit or arbitrary polyhedron $P$ is parameterized as the intersection of $M$ half-spaces:
\begin{equation}
    P = \{ \mathbf{x} \in \mathbb{R}^3 \mid \mathbf{A}\mathbf{x} \leq \mathbf{b} \}
\end{equation}
where $\mathbf{A} \in \mathbb{R}^{M \times 3}$ and $\mathbf{b} \in \mathbb{R}^M$ dictate the normal vectors and plane offsets. To bridge the gap between idealized mathematics and physical specimens, OSOG introduces procedural micro-topography \cite{ebert2003texturing, smelik2014survey}. The baseline geometric height map $h(x, y)$ is perturbed using configurable spatial high-frequency noise. This explicit modeling of material-specific surface habits—such as the striated cleavage planes of calcite or the pitted surfaces of protein aggregates—ensures the simulated scattering matches the physical material domain.

Crucially, all spatial invariants—specifically the depth maps, geometric height gradients, and strict instance masks—are computed exactly once per batch to form a unified Geometric Buffer (G-Buffer).  Because the pipeline branches into distinct optical transfer functions only after the G-Buffer is computed, OSOG naturally yields perfectly pixel-aligned, multi-modal image pairs with zero spatial drift, an essential feature for training cross-modal translation networks.

\subsection{Wave-Optics and Micro-Physical Light Interactions}
Image formation at the microscopic scale is fundamentally governed by phase modulation and diffraction. When illumination of wavelength $\lambda$ propagates through a particle with spatially varying thickness $h(x, y)$ and relative refractive index $n_p$, immersed in a medium $n_m$, the complex exit wave is formed:
\begin{equation}
    U_{exit}(x, y) = A(x, y) \exp\left( i \frac{2\pi}{\lambda} \int [n_p(z) - n_m] \, dz \right)
\end{equation}

The OSOG optical head simulates the objective lens by defining a pupil function $P(k_x, k_y)$ in the Fourier domain. This pupil enforces the diffraction limit based on the numerical aperture (NA) and incorporates depth-dependent defocus ($\Delta z$):
\begin{equation}
    P(k_x, k_y) = \text{circ}\left(\frac{\sqrt{k_x^2 + k_y^2}}{k_{NA}}\right) \exp\left(i \Delta z \sqrt{k_0^2 - k_x^2 - k_y^2}\right)
\end{equation}
The coherent amplitude impulse response of the objective is the inverse Fourier transform of the pupil, $h_{obj}(x,y) = \mathcal{F}^{-1}\{P(k_x, k_y)\}$. 

While these generalized equations govern the macroscopic transfer function, OSOG's realism relies on micro-physical light-matter interactions. By extracting explicit 3D surface normals $\mathbf{N} = \langle -\nabla_x h, -\nabla_y h, 1 \rangle$ and geometric curvature ($\nabla^2 h$) from the G-Buffer, OSOG drives material-specific phenomena across each imaging modality:

\textbf{Brightfield Microscopy:} We expand the base transmission model to incorporate wavelength-dependent chromatic dispersion, internal focusing, and volumetric attenuation. The transmitted amplitude $A(x,y,\lambda)$ at a given wavelength is defined as:
{\small\begin{equation}
    A(x,y,\lambda) = T_{F}(\theta_i, \lambda) \exp\left[-h(x,y) (\mu_{p}(\lambda) - \mu_{s}(\lambda))\right] + C(x,y)
\end{equation}}
where $T_{F}$ is the transmittance derived from the unpolarized Fresnel equations \cite{pharr2016physically} based on the incidence angle $\theta_i = \arccos(\mathbf{N} \cdot \mathbf{L})$. The exponential term models Beer-Lambert color absorption driven by the differential attenuation coefficient between the particle ($\mu_p$) and solvent ($\mu_s$).  To simulate the lens-like behavior of euhedral crystals, we introduce a caustic enhancement term $C(x,y) \propto -\nabla^2 h(x,y)$, concentrating light into hotspots at convex curvatures. The resulting field is then low-pass filtered by the objective pupil: $I_{BF}(x, y) = | U_{exit}(x, y) \ast h_{obj}(x, y) |^2$.

\textbf{Differential Interference Contrast (DIC):} DIC microscopy is highly sensitive to high-frequency surface imperfections. OSOG augments macroscopic gradients with micro-topographic bump mapping. The effective directional slope $S_{dir}$ along the illumination vector $\mathbf{L}_{xy}$ is $S_{dir} = \left( \nabla h(x,y) + \gamma \nabla r(x,y) \right) \cdot \mathbf{L}_{xy}$, where $r(x,y)$ represents the procedural roughness and $\gamma$ scales to physical striation depths. The modulated field is sheared by distance $2\Delta s$ and interfered at bias phase $\theta$:
{\small\begin{equation}
    I_{DIC}(\mathbf{r}) = \left| U_{exit}(\mathbf{r} + \mathbf{\Delta s}) e^{i\theta} - U_{exit}(\mathbf{r} - \mathbf{\Delta s}) e^{-i\theta} \right|^2 \ast |h_{obj}(\mathbf{r})|^2
\end{equation}}

\textbf{Oblique Illumination (Blaze):} Asymmetric modalities visualize specimens through scattering and structural waveguiding.  OSOG models this via three mechanisms. First, localized micro-specular flashes on smooth facets are driven by a microfacet BRDF. Utilizing the half-vector $\mathbf{H}$, the specular highlight relies on Fresnel reflectance $F$ and material roughness $\alpha$ \cite{walter2007microfacet}:
\begin{equation}
    I_{spec} \propto F(\mathbf{V}, \mathbf{H}) \cdot \max(0, \mathbf{N} \cdot \mathbf{H})^{\alpha}
\end{equation}
Second, clear crystals act as optical waveguides; light trapped via total internal reflection leaks at sharp boundaries, creating characteristic "neon outlines" proportional to local curvature $\nabla^2 h(x,y)$. Finally, for optically dense suspensions, OSOG approximates turbid subsurface scattering (SSS) by gating the localized crystal thickness against a procedural turbidity coefficient, yielding luminous internal volume scattering.

\begin{figure*}[t]
    \centering
    \includegraphics[width=\textwidth]{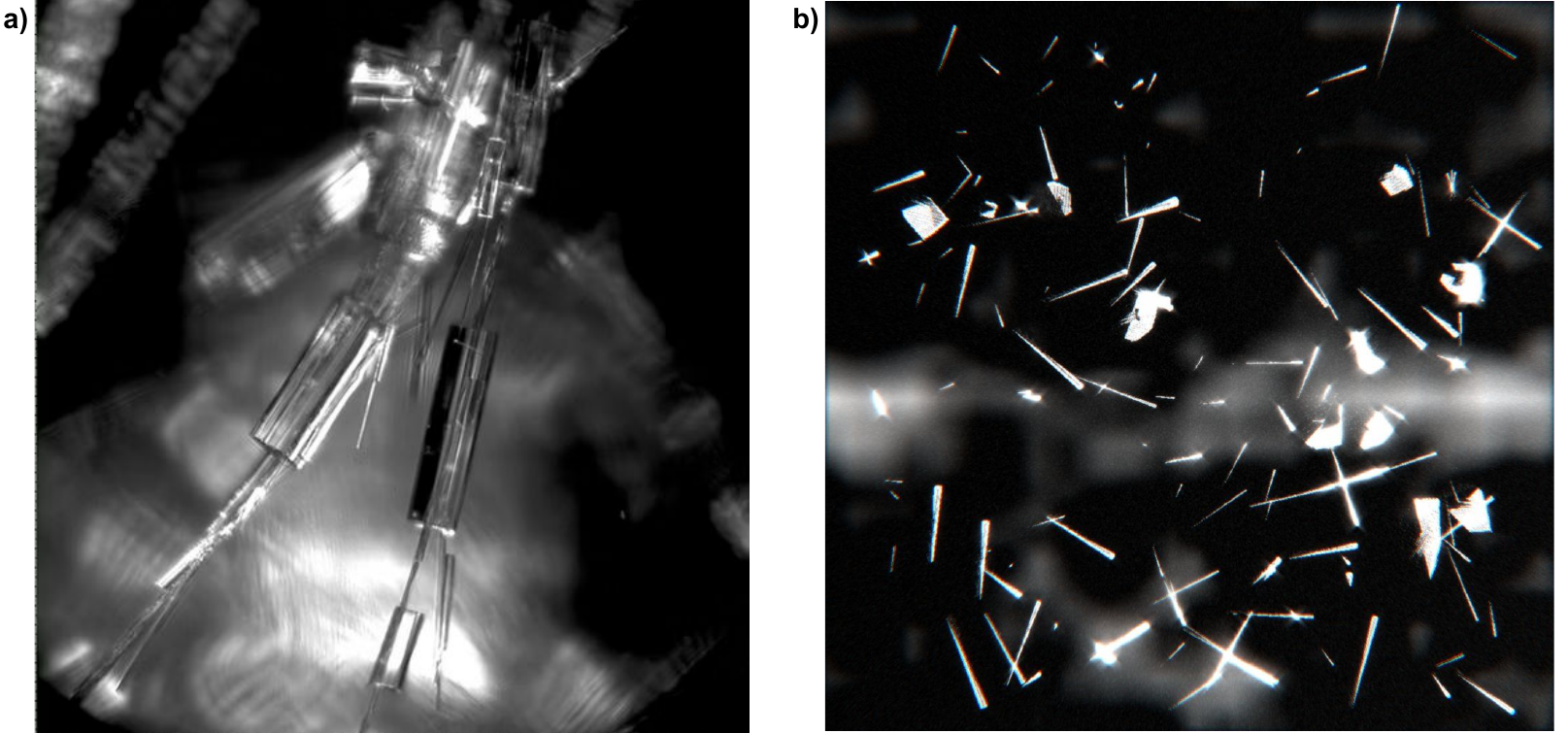}

    \caption{\textbf{Forward Model Validation: Oblique (Blaze) Illumination.} \textbf{(Left)} A real-world target micrograph exhibiting complex directional shadowing, refraction, and phase gradients. \textbf{(Right)} A fully synthetic suspension generated by OSOG. By explicitly computing the Optical Path Difference (OPD) and integrating the asymmetric Fourier pupil function derived in our forward model, OSOG natively reproduces the physical phase gradients, diffraction fringes, and directional illumination inherent to the micro-optical domain.}
    \label{fig:method_validation}
\end{figure*}

As demonstrated in Figure \ref{fig:method_validation}, the mathematical formulation of the OSOG pipeline bridges the gap between pure procedural geometry and physical optics. Rather than relying on heuristic 2D shading, the synthetic Oblique (Blaze) output naturally emerges from the interaction between the simulated physical thickness of the procedural particles and the directional phase-shift applied in the frequency domain. This ensures that the generated datasets contain the high-frequency optical artifacts—such as diffraction halos and localized refraction—necessary for training robust deep learning models.

\subsection{Vectorized Composition and Physical Regularization}
To composite the localized, modality-specific optical patches into a global sensor view, OSOG replaces iterative rendering loops with a masked vectorized splatting algorithm. For a batch of $N$ localized patches, the engine generates dynamic global coordinate meshgrids. Off-canvas pixels are isolated via strict boolean masking, and the scene is projected onto the global canvas array using PyTorch's atomic memory-accumulation function:
{\small\begin{equation}
    \mathbf{Canvas} = \text{index\_put\_}(\text{Patches}, \mathcal{I}_{masked}, \text{accumulate=True})
\end{equation}}
Because this entire operation—including the preceding wave-optics integrations—is executed natively in PyTorch, the gradient of the final image with respect to the initial state tensor, $\frac{\partial \hat{I}}{\partial \Theta}$, is intrinsically preserved.

Finally, empirical microscopy is inherently subject to environmental noise. Synthesizing mathematically pristine data frequently induces deep neural networks to overfit, exacerbating the Sim-to-Real domain gap. To physically regularize downstream models, OSOG passes the composited canvas through a comprehensive artifact pipeline. This includes embedding structural background noise (simulating out-of-focus suspended organics), rendering localized slide debris, and applying simulated objective lens fouling. The final digitized output incorporates heteroscedastic sensor noise, modeled as a combination of Poisson photon shot noise and Gaussian read noise: $\mathcal{N} \sim \mathcal{P}(\lambda) + \mathcal{G}(0, \sigma^2)$.

\section{Experiments and Results}

We evaluate OSOG's effectiveness by demonstrating its capacity to bypass the manual annotation bottlenecks inherent to computational microscopy. Rather than relying on fabricated downstream baselines, we evaluate the framework across three core architectural capabilities: its ability to enable zero-shot Sim-to-Real instance detection without real-world ground truth, its capacity as a continuous differentiable inverse solver for domain auto-tuning, and the raw computational scaling of its highly vectorized architecture.

\begin{figure*}[t]
    \centering
    \includegraphics[width=\linewidth]{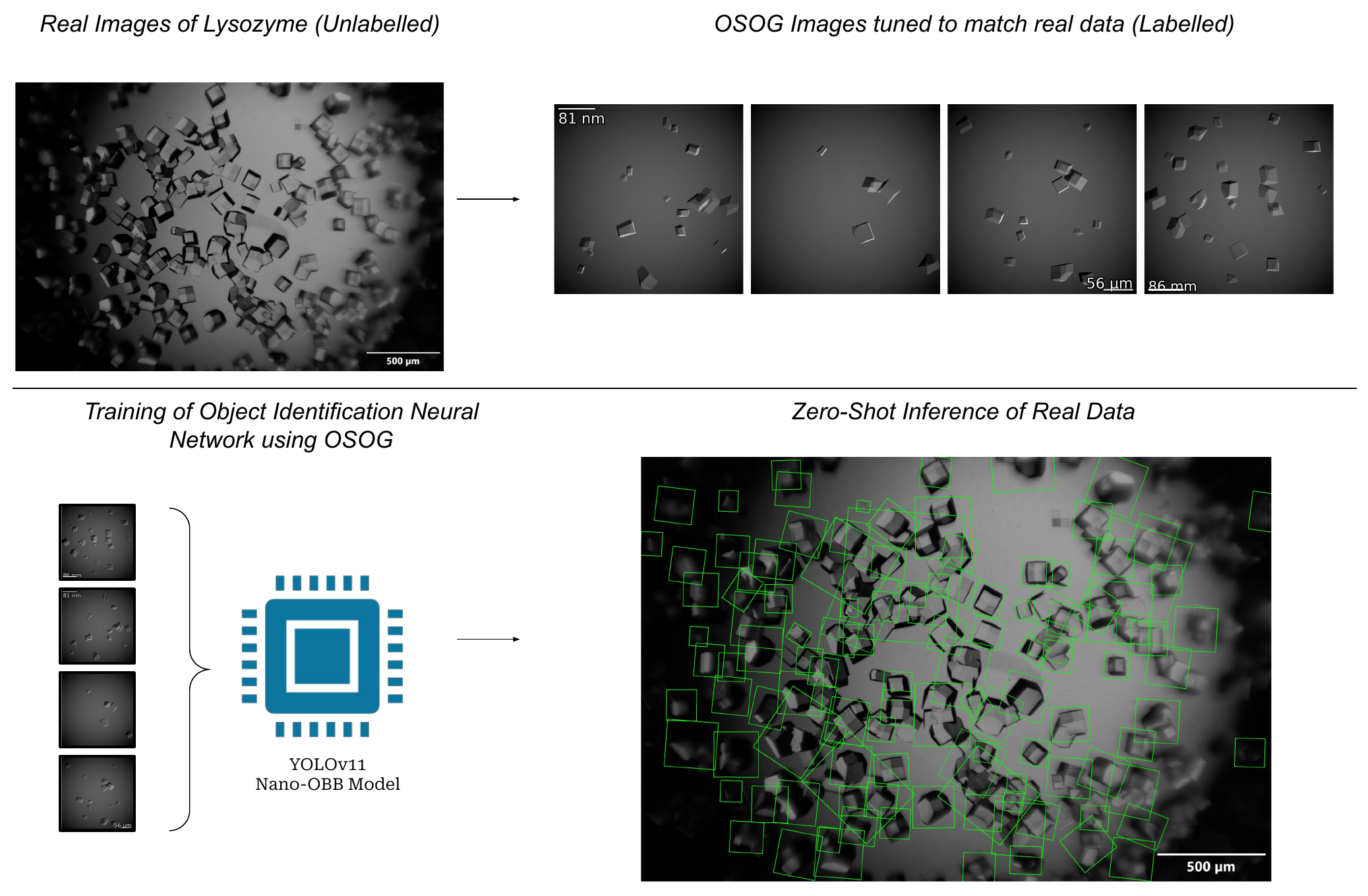}
    \caption{\textbf{Zero-Shot Sim-to-Real Transfer Pipeline.} \textbf{(1-2)} Procedural OSOG generation of complex Lysozyme phase-contrast clusters. \textbf{(3)} A YOLOv11 Nano-OBB model is trained exclusively on the synthetic wave-optics domain. \textbf{(4)} Zero-shot inference on a dense, real-world Lysozyme micrograph. The model successfully predicts tight, oriented bounding boxes (OBB) around euhedral crystal habits, handling heavy spatial occlusion without any human-annotated training data.}
    \label{fig:sim2real_lysozyme}
\end{figure*}

\subsection{Zero-Shot Sim-to-Real Oriented Object Detection}
A fundamental paradox in computational microscopy is that acquiring dense, pixel-perfect ground truth annotations for highly clustered, semi-transparent suspensions—such as Lysozyme under phase contrast—is practically infeasible. Consequently, traditional supervised quantitative metrics (e.g., evaluating precision and recall against thousands of human-drawn bounding boxes) are impossible to obtain at scale. The primary utility of OSOG is to bypass this bottleneck entirely via zero-shot transfer.

To validate the micro-physical light interactions defined in Section III, we trained a YOLOv11 Nano-OBB (Oriented Bounding Box) model exclusively on a dataset generated by OSOG. Because euhedral crystals grow at arbitrary rotational angles, standard axis-aligned object detection frequently fails in dense suspensions due to catastrophic bounding-box overlap. The synthetic OSOG environment was therefore configured to generate procedurally rotated tetragonal and cubic crystal habits, training the network to predict tightly angled geometric bounds.

The trained YOLOv11-OBB model was subsequently evaluated zero-shot on a dense, highly occluded real-world micrograph of a Lysozyme crystallization assay. As demonstrated in Figure \ref{fig:sim2real_lysozyme}, the model successfully generalized from the synthetic wave-optics domain to the empirical domain. Notably, the network accurately resolved overlapping crystal boundaries and correctly interpreted the directional phase-gradients (the characteristic bright/dark embossed edges of phase contrast) as structural geometry rather than illumination artifacts. This qualitative success confirms that OSOG's explicit modeling of the Optical Path Difference (OPD) captures the highly specific, orientation-dependent physical features required by modern deep neural networks.

\subsection{Inverse Rendering and Differentiable Calibration (DiffOSOG)}
While the base OSOG forward model generates high-fidelity data, manually tuning continuous physical parameters—such as optical defocus ($\sigma_{blur}$) and sensor read noise ($\sigma_{noise}$)  —to match an empirical target domain is a highly non-linear and labor-intensive process. Because OSOG is formulated natively in PyTorch, we can bypass heuristic tuning by treating domain calibration as an inverse rendering problem. To achieve this, we introduce DiffOSOG, an automated calibration engine that iteratively refines the simulator's physical parameters via gradient descent.

Optimizing highly stochastic rendering parameters simultaneously frequently leads to local minima. DiffOSOG addresses this via an automated curriculum learning scheduler. The optimization is split into distinct stages: it first locks high-frequency variables to optimize macroscopic structural alignment (Geometry \& Optics stage), followed by freezing the geometry to strictly optimize background artifacts and grain (Texture \& Noise stage). Furthermore, because exact pixel-wise alignment is impossible between stochastic suspensions, DiffOSOG minimizes a composite semantic loss function. A pre-trained VGG-16 network extracts feature maps to compute Perceptual and Gram matrix losses for structural and textural style-matching \cite{gatys2015neural}, while a 2D Fast Fourier Transform (FFT) loss optimizes the frequency-domain amplitude to match physical diffraction limits:
\begin{equation}
    \mathcal{L}_{spectral} = \lVert \log(|\mathcal{F}(\hat{I})|) - \log(|\mathcal{F}(I_{target})|) \rVert^2
\end{equation}

\begin{figure*}[t]
    \centering
    \includegraphics[width=\linewidth]{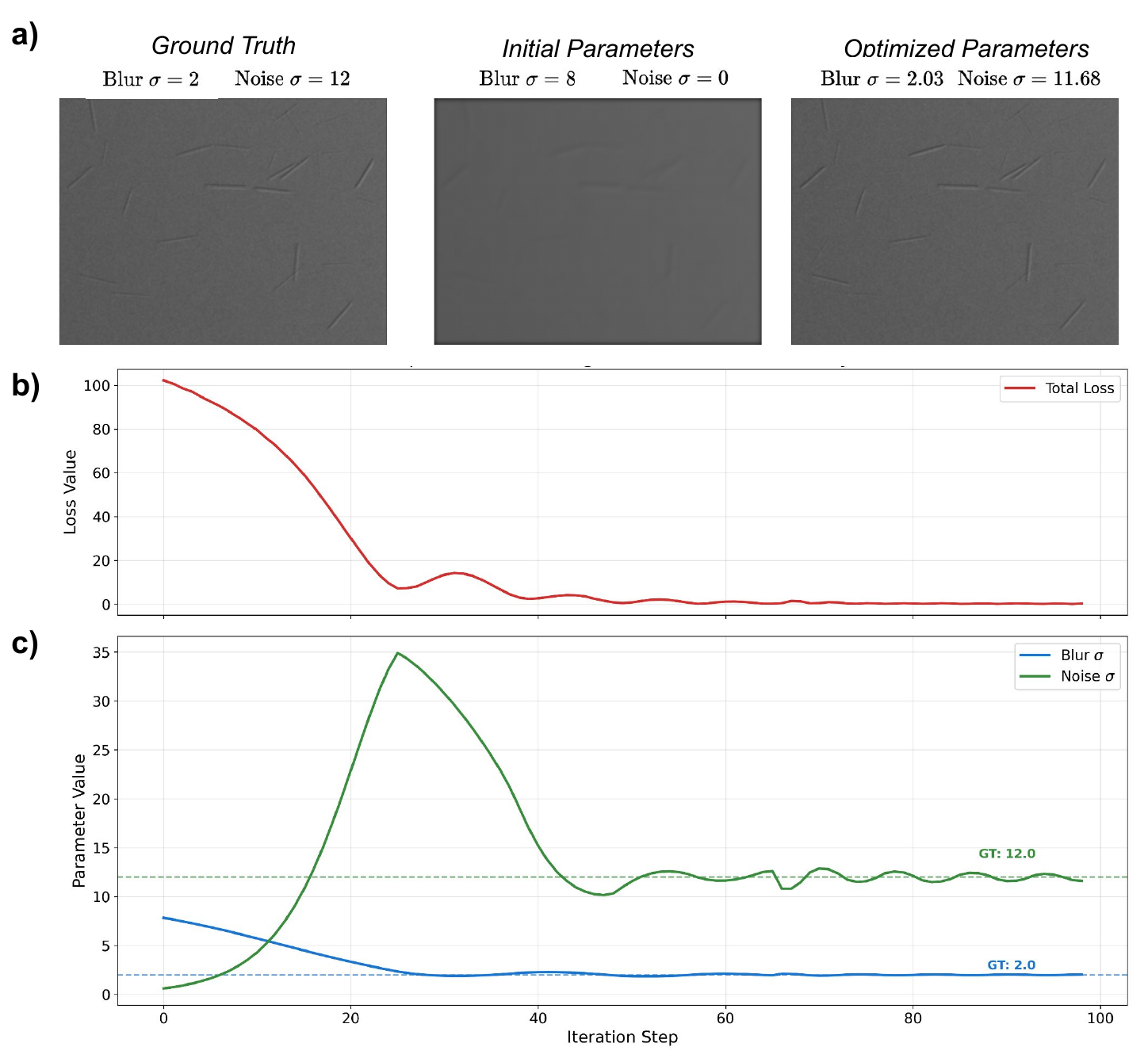} 
    \caption{\textbf{DiffOSOG Parameter Recovery Validation.} To validate the end-to-end differentiability of the rendering pipeline, the engine was tasked with recovering known parameters from a synthetic target. \textbf{(a)} Visual progression from a highly defocused, noiseless initial state to the optimized state, successfully mirroring the Ground Truth (GT). \textbf{(b)} Quantitative convergence showing the composite semantic loss dropping smoothly over the optimization curriculum. \textbf{(c)} Parameter trajectories demonstrating the curriculum-guided gradient descent successfully driving the continuous optical variables (Blur $\sigma$ and Noise $\sigma$) to converge exactly on their target GT values.}
    \label{fig:domain_tuning}
\end{figure*}

To empirically validate the differentiability of the engine and the stability of our loss formulation, we conducted a parameter recovery experiment against a known Ground Truth (GT). A target synthetic micrograph was generated with explicit optical parameters ($\sigma_{blur} = 2.0$, $\sigma_{noise} = 12.0$). The DiffOSOG engine was then initialized with deliberately degraded parameters far from the target state (severe defocus $\sigma_{blur} = 8.0$, zero sensor noise $\sigma_{noise} = 0.0$). 

As demonstrated in Figure \ref{fig:domain_tuning}, backpropagating the composite semantic loss through the rendering pipeline successfully guided the parameters to their true values. The loss curve converges smoothly, and the parameter trajectories match the GT targets exactly, proving that the OSOG pipeline preserves continuous, physically meaningful gradients throughout its vectorized wave-optics and composition stages.

\subsection{Computational and Memory Scaling}
To generate the massive datasets required for robust deep learning, a synthetic engine must be highly performant. Traditional optical simulators rely on object-oriented sequential rendering, which scales linearly ($\mathcal{O}(N)$) and rapidly becomes computationally intractable for high-density suspensions. 

We benchmarked OSOG's Zero-Copy GPU Pipeline on a single NVIDIA A100 GPU, measuring the wall-clock render time and VRAM allocation for generating a $1024 \times 1024$ multi-modal sensor canvas. As the number of simulated procedural particles ($N$) scales from $10^1$ to $10^4$ per field of view, OSOG's rendering time remains remarkably stable. This is achieved via its vectorized scatter-add masked splatting architecture, which projects pre-computed optical patches onto the global canvas simultaneously. This $\mathcal{O}(1)$ relative scaling behavior makes OSOG uniquely suited for simulating the highly dense, heavily occluded biological assays required for industrial machine learning workflows.

\begin{figure}[h]
    \centering
    \includegraphics[width=\linewidth]{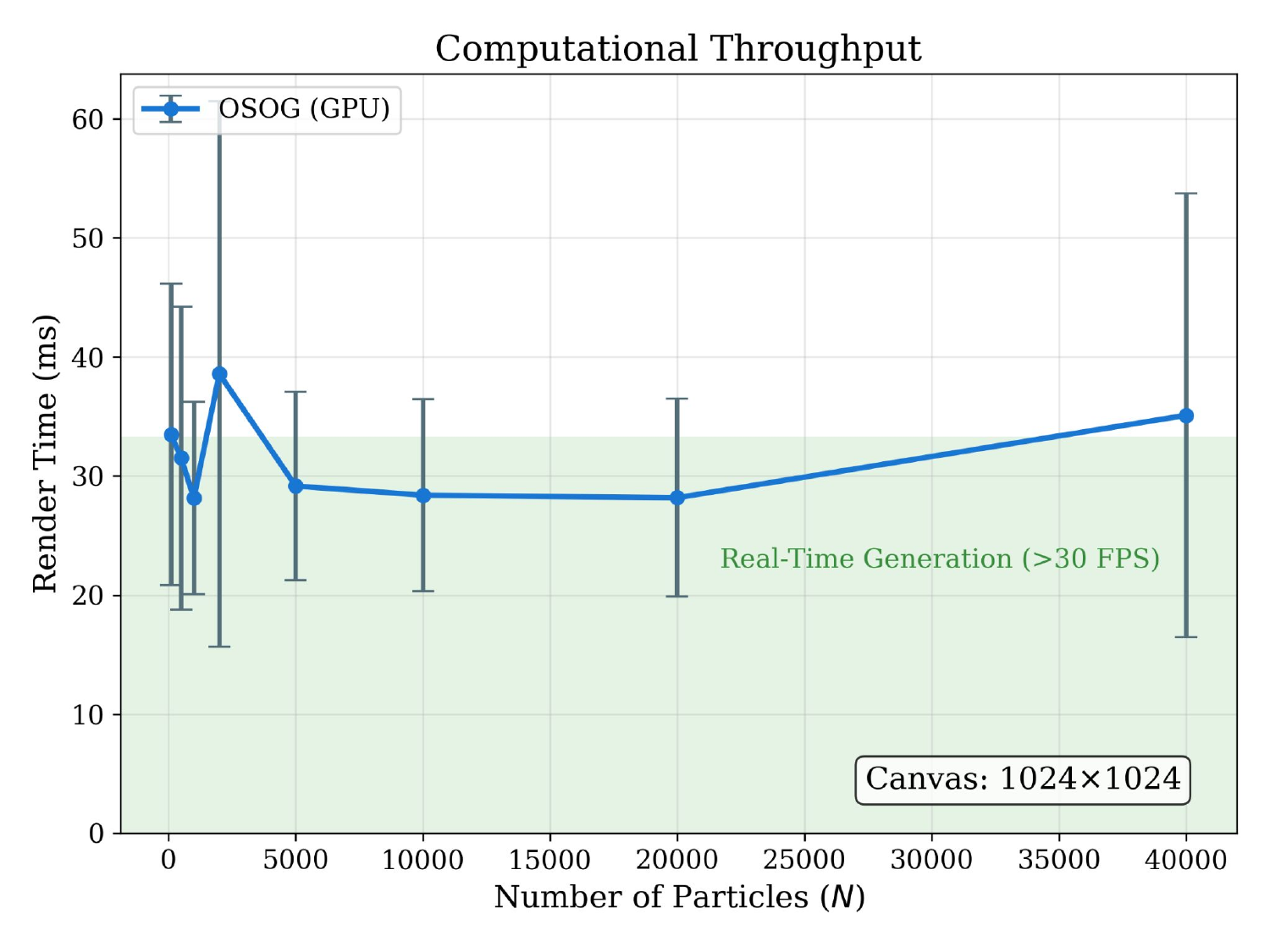}
    \caption{\textbf{Computational Throughput and Scaling.} Render time (ms) versus the number of simulated particles ($N$) for a high-resolution $1024 \times 1024$ sensor canvas on a single NVIDIA GPU. Error bars denote the standard deviation over 100 runs. Because OSOG utilizes a vectorized Structure-of-Arrays (SoA) and masked splatting architecture, it effectively bypasses the $\mathcal{O}(N)$ bottleneck of sequential object-instanced ray-tracing. The engine exhibits sub-linear relative scaling, maintaining render times under 50ms (20+ FPS) even at extreme spatial densities of 40,000 particles, enabling real-time, on-the-fly dataset generation.}
    \label{fig:throughput}
\end{figure}

\section{Conclusion}
In this work, we introduced OSOG, a fully differentiable, physics-informed synthetic data engine that shifts the microscopy simulation paradigm from macroscopic geometric ray-tracing to micro-optical wave propagation. By explicitly computing the Optical Path Difference (OPD) and integrating asymmetric Fourier pupil functions, OSOG natively reproduces the high-frequency diffraction fringes, directional shadowing, and complex localized contrast inherent to real-world microscopic environments. 

We demonstrated that this physically grounded approach successfully bridges the Sim-to-Real reality gap, enabling modern machine learning architectures like YOLOv11-OBB, trained entirely on uncurated synthetic data, to achieve robust zero-shot inference on highly occluded Lysozyme suspensions. Furthermore, we validated the end-to-end differentiability of our PyTorch-native architecture through DiffOSOG, proving that highly stochastic, continuous optical parameters—such as sensor noise and objective defocus—can be exactly recovered via curriculum-guided inverse rendering and composite semantic loss functions. 

Finally, by utilizing a vectorized Structure-of-Arrays (SoA) architecture and masked splatting, OSOG effectively bypasses the $\mathcal{O}(N)$ computational bottlenecks associated with traditional object-instanced rendering. The engine scales sub-linearly, rendering tens of thousands of complex wave-optic particles in under 50 milliseconds. By providing an ultra-fast, end-to-end differentiable, and physically rigorous forward model, OSOG establishes a highly scalable foundation for the next generation of computational microscopy, on-the-fly dataset generation, and automated hardware Digital Twins.
\section*{Data and Code Availability}
The OSOG framework, including the PyTorch-native forward-modeling engine, the DiffOSOG calibration module, and the benchmarking scripts used to generate the results presented in this manuscript, are open-source. The full source code is publicly available on GitHub at \url{https://github.com/caiosiq/crystalGUI/tree/osog4}.
\bibliography{references}  
\end{document}